\pdfoutput=1

\documentclass[11pt]{article}

\usepackage[final]{acl}

\usepackage{times}
\usepackage{latexsym}

\usepackage[T1]{fontenc}

\usepackage[utf8]{inputenc}

\usepackage{microtype}

\usepackage{inconsolata}

\usepackage{graphicx}

\usepackage{booktabs}
\usepackage{amsmath}
\usepackage{multirow}
\usepackage{amssymb}
\usepackage{xcolor,pifont}
\usepackage{placeins}

\title{
A Close Look at Decomposition-based XAI-Methods \\ for Transformer Language Models
}

\author{
 \textbf{Leila Arras\textsuperscript{1,2}},
  \textbf{Bruno Puri\textsuperscript{1,3}},
  \textbf{Patrick Kahardipraja\textsuperscript{1}},\\
  \textbf{Sebastian Lapuschkin\textsuperscript{1}},
  \textbf{and Wojciech Samek\textsuperscript{1,2,3}} \\
  \textsuperscript{1}Department of Artificial Intelligence, Fraunhofer Heinrich Hertz Institute, Berlin, Germany\\
  \textsuperscript{2}BIFOLD - Berlin Institute for the Foundations of Learning and Data, Berlin, Germany \\
  \textsuperscript{3}Department of Electrical Engineering and Computer Science, Technische Universität Berlin, Berlin, Germany
\\
{\texttt{\{leila.arras, sebastian.lapuschkin, wojciech.samek\}@hhi.fraunhofer.de}}} 

\begin{document}
\maketitle
\begin{abstract}
Various XAI attribution methods have been recently proposed   for the transformer 
architecture, allowing for insights into the decision-making process of large language models by assigning importance scores to input tokens and intermediate representations. One class of methods that seems very promising in this direction 
includes \textit{decomposition}-based approaches, i.e., XAI-methods that redistribute the model's prediction \textit{logit} through the network,  
as this value is directly related to the prediction. In the previous literature we note though that two prominent methods of this category, 
namely ALTI-Logit and LRP, have not yet been  analyzed in juxtaposition and 
hence we propose to close this gap by conducting a careful quantitative evaluation w.r.t.\ ground truth annotations on a subject-verb agreement task, 
as well as various qualitative inspections, using BERT, GPT-2 and \mbox{LLaMA-3} as a testbed. 
Along the way we compare and extend the %
ALTI-Logit and LRP methods, including the recently proposed AttnLRP variant, from an algorithmic and implementation perspective. We further incorporate in our benchmark two widely-used gradient-based attribution techniques. 
Finally, we make our carefullly constructed benchmark dataset for evaluating attributions on language models, as well as our code\footnote{Link will be made available upon paper acceptance.}, 
publicly available in order to foster evaluation 
of XAI-methods on a well-defined common ground.
\end{abstract}


\begin{figure}[ht]
    \centering
    \includegraphics[trim={0cm 0cm 0cm -0.25cm},clip,page=1, scale=0.65]{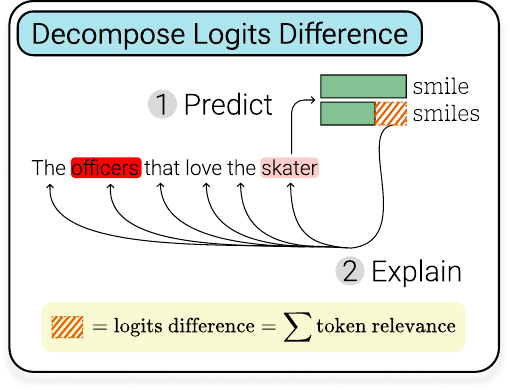}
    \includegraphics[trim={0cm 0cm 0cm 0cm}, clip, page=2, scale=0.65]{figures/benchmark_pipeline.pdf}
    \caption{Our XAI evaluation pipeline using subject-verb agreement: 1) Predict the logits difference for the two verb forms, 2) Explain the logits difference by generating a token-level relevance heatmap for each XAI-method (for decomposition-based XAI-methods the relevances sum up to the logits difference), 3) Evaluate the heatmaps w.r.t. ground truth linguistic evidence (i.e., the verb's subject) by computing various relevance accuracy metrics (such as, e.g., the fraction of positive relevance falling inside the GT).}
    \label{fig:benchmarkpipeline}
\end{figure}

\section{Introduction \& Background}

\subsection{Interpretability of Transformers}

Many approaches have been explored to shed light on how Transformer models process language. BERTology works \cite{rogers-etal-2020-primer} primarily employ probes \cite{gupta-etal-2015-distributional, kohn-2015-whats,alain2016understandingintermediatelayersusing} 
to analyze what information the model's internal representations encode, which can range from linguistic properties to factual and world knowledge \citep[{i.a.}]{clark-etal-2019-bert, hewitt-manning-2019-structural, liu-etal-2019-linguistic, petroni-etal-2019-language, tenney-etal-2019-bert, cui-etal-2021-commonsense}. However, probing itself is not without limitations, as it is correlational in nature and requires careful probes selection and interpretation \cite{hewitt-liang-2019-designing, belinkov-2022-probing}.

This spurs another line of inquiry asking how information is actually being used via causal intervention \cite{judea2001direct, vig2020investigating, geiger2021causal}. \citet{elazar-etal-2021-amnesic} propose amnesic probing that ablates certain linguistic properties such as part-of-speech from models' representation to see how it affects actual predictions. Similarly, \citet{meng2022locating} analyse how LLMs store and recall factual information by intervening on weights and hidden representations. Causal methods allow the isolation of subgraphs of neural networks that are responsible for certain tasks such as indirect object identification \cite{wang2023interpretability} and induction heads \citep{olsson2022incontextlearninginductionheads}, although the process itself relies on a non-trivial amount of manual labor. Recent efforts such as ACDC \citep{conmy2023acdc} attempt to alleviate this issue, yet still can miss some nodes that are supposed to be part of the subgraph.

Our focus lies on attribution methods, specifically those that \textit{decompose} the prediction's logit throughout the entire network and do not only consider parts of the Transformer model such as  MLPs \cite{geva-etal-2021-transformer, geva-etal-2022-transformer} or attention modules \cite{ abnar-zuidema-2020-quantifying, kobayashi-etal-2020-attention}. These approaches are able to measure causal properties \citep{geiger2021causal}, allowing for the identification and localization of features playing an important role during inference in a more scalable manner, and simultaneously enable inspection of information encoded in the model \cite{Achtibat2023, ferrando-voita-2024-information}.

\subsection{Evaluation of Attributions}

A common way to evaluate attributions is to systematically perturb parts of the models' inputs according to their relevance and then measure the resulting changes in the output, the higher the change the more accurate the attribution.
Such an approach has been initially proposed as pixel-perturbation in the computer vision domain \cite{bach2015lrp, samek2016evaluating}, and was later extended to words and tokens in NLP \cite{arras-etal-2016-explaining,arras-etal-2019-evaluating,deyoung-etal-2020-eraser,atanasova-etal-2020-diagnostic}.

Another direction is to use syntactic tasks to evaluate attributions, such as subject-verb agreement \cite{poerner-etal-2018-evaluating,Linzen_TACL2016}, since those tasks typically allow for the creation of ground truth annotations.
Although such an approach has been quite popular, to the best of our knowledge there exist no properly constructed and publicly available benchmark dataset using subject-verb agreement on real-world natural language data; existing benchmarks often consist of artificially created data using short and simple sentence templates, such as done for instance in BLiMP and CausalGym \cite{warstadt-etal-2020-blimp-benchmark,arora-etal-2024-causalgym}.

Besides automatic evaluation, user studies were also widely used to evaluate explanations \cite{doshivelez2017rigorousscienceinterpretablemachine, lipton2018mythos, hase-bansal-2020-evaluating,jacovi-goldberg-2020-towards}.

Given this context, our contributions in the present work are the following:
\texttt{(a)} Analyze and compare decomposition-based attribution methods which were not yet compared to one another (namely ALTI-Logit and LRP/AttnLRP); 
\texttt{(b)} Generate and release a ground truth annotated real-world dataset for evaluating attributions on Language Models using a subject-verb agreement task;
\texttt{(c)} Extend the ALTI-Logit XAI-method to the Llama model family;
\texttt{(d)} Propose a novel fast and simple method to implement the AttnLRP XAI-method based on a \textit{modified} Gradient$\times$Input strategy, as well as provide a complete set of proofs 
        to justify this approach for both  XAI-methods LRP and AttnLRP.

\section{XAI-Methods Strategy}

Let us first introduce some notations that will help us analyze and compare the strategy of the considered XAI-methods. Let $x_t^l$ be the token representation for timestep $t$ and layer $l$, and $R_t^l$ the corresponding relevance\footnote{In this work we use the terms relevance, contribution, attribution and importance score interchangeably.} for this token. Accordingly $R_t^0$ represents the relevance of the input token for timestep $t$. Let $\mathbf{U}$ be the output embedding matrix, and $U_w$ the column vector for the predicted token $w$. 
Hence, the language model's prediction logit for predicting token $w$ at timestep $T$ is\footnote{Here we assume an Autoregressive Language Model, with $T$ being the input length, but all considered XAI-methods are in principle applicable to Masked Language Models as well.}: $\text{logit}_w = x_T^L \cdot U_w$, with $L$ being the number of layers of the model.
A property which is common to all decomposition-based XAI-methods is that the $\text{logit}_w$ is decomposed additively into contributions of model components (token, neuron, head or layer), or in other words, the contributions of model components sum up to the value $\text{logit}_w$.

\subsection{ALTI-Logit}
ALTI-Logit is a recently proposed state-of-the-art decomposition-based approach for Transformer Language Models  proposed by \citet{ferrando-etal-2023-explaining}.
Its main idea is to additively decompose the final layer's token representation $x_T^L$  used to compute the prediction logit (i.e., the penultimate vector ahead of the output embedding layer $\mathbf{U}$) into layer-wise contributions of the outputs of each MLP and MHA block\footnote{We refer to MLP as Multi-Layer Perceptron, and MHA as Multi-Head Attention, representing the two main components of the Transformer architecture.}, by following the residual connections of the model \cite{elhage2021mathematical}.
While the contributions of the MLP blocks are not decomposed further backward, the contributions of the MHA blocks get further broken down into contributions of their respective input token representations, similarly to attention decomposition from \citet{kobayashi-etal-2021-incorporating}.
The latter is achieved by linearizing the MHA-block by viewing the attention weight matrix as a constant, as well as treating the standard deviation within the normalization layer as a constant, similarly to how  Layer-wise Relevance Propagation (LRP) was previously extended to Transformers \cite{pmlr-v162-ali22a}.
Lastly, in order to account for the mixing of information across multiple layers, a token-level contribution matrix is built within each MHA block by considering the contributions of the MHA's transformed vectors to the MHA's output vector (as was done in the ALTI method by \citet{ferrando-etal-2022-measuring}), and the resulting matrices are multiplied across layers to finally obtain an ALTI-Logit contribution for each input token. Overall the decomposition property of ALTI-Logit can be summarized as follows\footnote{Ignoring contributions from model biases for simplicity of notation.}:
$\sum_t R_t^0 + \sum_{l} \tilde{R}_T^l =\text{logit}_w$, where $R_t^0$ is the input token contribution for each timestep $t$ in the input sequence resulting from the MHA blocks and aggregated over all layers, while $\tilde{R}_T^l$ is the contribution of the output of each MLP block for the given prediction timestep $T$ and layer $l$, since ALTI-Logit assumes that there is no mixing of information across timesteps resulting from MLP blocks.

In practice, the official implementation of ALTI-Logit\footnote{\url{https://github.com/mt-upc/logit-explanations}} by \citet{ferrando-etal-2023-explaining} requires the computation of a second, carefully designed forward pass through the model (using attention matrices, as well as weight parameters from various intermediate layers), after having run a first standard forward pass through the model during which the inputs and outputs of hidden layers are collected via hooks.
This dedicated forward pass in ALTI-Logit was so far derived for Pre-LayerNorm architectures\footnote{We refer to \textit{Pre}-LayerNorm to indicate that the normalization layer is located \textit{before} the self-attention computation (resp.\ the fully-connected layers) within the MHA (resp.\ MLP) blocks, as opposed to \textit{Post}-LayerNorm where the normalization happens \textit{after} them.} and Autoregressive Language Models,
and exclusively applied to the models GPT2 \cite{radford2019language}, OPT \cite{zhang2022optopenpretrainedtransformer} and BLOOM \cite{workshop2023bloom176bparameteropenaccessmultilingual}.

In  this work we extend the ALTI-Logit algorithm to the Llama model family \cite{touvron2023llama2openfoundation, grattafiori2024llama3herdmodels} by  adapting ALTI-Logit to handle grouped-query attention \cite{ainslie-etal-2023-gqa}, as well as RMSNorm normalization \cite{NEURIPS2019_1e8a1942}.
However, we refrain from adapting ALTI-Logit to the BERT model family, as this would require a substantial re-design of the algorithm to cope with Post-LayerNorm  architectures as well as Masked Language Modeling.

ALTI-Logit provides layer-wise token-level (as well as head-level) contributions to the prediction logit, and this method
(resp. its components Logit \cite{ferrando-etal-2023-explaining} and ALTI \cite{ferrando-etal-2022-measuring}) were previously evaluated against Erasure \cite{li2017understandingneuralnetworksrepresentation}, Gradient \cite{Simonyan:ICLR2014,li-etal-2016-visualizing}, Gradient$\times$Input \cite{Denil:2015,Shrikumar:arxiv2016}, Integrated Gradients \cite{Sundararajan:ICML2017}, Attention Rollout \cite{abnar-zuidema-2020-quantifying} and GlobEnc \cite{modarressi-etal-2022-globenc} explanations, where ALTI-Logit was shown to deliver the best results.

\subsection{Layer-wise Relevance Propagation}\label{sec:LRP}
Layer-wise Relevance Propagation (LRP) \cite{bach2015lrp} is an interpretability method based on a backward decomposition
following a layer-wise conservation principle. In other words, in each layer of the model the contributions of neurons sum up to the prediction logit. More precisely it holds\footnote{Here also ignoring the relevances assigned to model biases for simplicity.}: $\sum_t R_t^0 = \sum_t R_t^1 = \dots = \sum_t R_t^L = \text{logit}_w$. LRP was initially proposed for Convolutional Neural Networks \cite{bach2015lrp}, and later extended to other models such as Recurrent Networks \cite{arras-etal-2017-explaining,ArrXAI19}, Transformers \cite{pmlr-v162-ali22a,pmlr-v235-achtibat24a} and selective State Space Models \cite{NEURIPS2024_d6d0e41e}.

In practice, LRP can be implemented by applying dedicated LRP backward propagation rules for each type of layer occuring in the network, and that redistribute neuron relevances from upper layers to lower layers in a conservative manner \cite{Montavon:ExplAIBook2020}.

For a linear layer with forward pass equation $ z_j = \sum_i z_i w_{ij} + b_j$, and given the relevances of the output neurons $R_j$,  
the input neurons' relevances $R_i$ are computed through a summation of the form\footnote{This rule corresponds to the LRP-$\epsilon$ rule (with $\epsilon$ being a small numerical stabilizer) which was shown to work well in NLP. On computer vision models, in particular for convolutional layers, other rules have be shown to be more adequate \cite{Montavon:ExplAIBook2020,ARRAS202214,9206975}}:
$ R_{i} =  \sum_j \tfrac{z_i \cdot w_{ij}}{z_j  \;  + \; \epsilon \cdot {\text{sign}}(z_j)} \cdot R_j \,$, hence their relevances are proportional to their forward pass contributions.
For an element-wise activation layer of the form $z_j = g(z_i)$, with $g$ being a non-linear activation function, the relevance $R_j$ is redistributed  backward using the identity rule, thus $R_i = R_j$.
In order to extend LRP to Transformer models, it is required to design new rules to propagate the relevance backward through two further non-linearities typical  to the models' architecture: product layers (occurring for instance in the product between attention weights and value vectors inside the MHA), and normalization layers (LayerNorm or RMSNorm).
To this end \citet{pmlr-v162-ali22a} propose to view the attention weights as a constant, which is equivalent to using the signal-take-all LRP redistribution rule for products which was previously proposed for extending LRP to Recurrent Neural Networks \cite{arras-etal-2017-explaining,ArrXAI19}.
For the normalization layers, \citet{pmlr-v162-ali22a} propose to treat the standard deviation as a constant. In practice, these two rules can be implemented by treating the previous non-linearities as linear layers for LRP (see Appendix~\ref{appendix:LRPx_proofs} for more details). 

While the LRP extension to Transformers has been proposed in \citet{pmlr-v162-ali22a}, early implementations of LRP on Transformers  \citet{pmlr-v162-ali22a,eberle-etal-2022-transformer} omit the redistribution of relevance through MLP blocks (more particularly through its element-wise activation layer), and were only utilizing LRP rules inside MHA blocks. To the best of our knowledge the first LRP implementation applied to a complete Transformer architecture was provided  by \citet{eberle-etal-2023-rather}. 
\citet{pmlr-v162-ali22a} evaluated LRP against various attention-based XAI-methods \cite{abnar-zuidema-2020-quantifying,sood-etal-2020-interpreting,9710570}, as well as Gradient$\times$Input \cite{Denil:2015,Shrikumar:arxiv2016}, and LRP was shown to deliver the best results.

\subsection{AttnLRP}\label{sec:AttnLRP}
AttnLRP is a novel variant of LRP \cite{pmlr-v235-achtibat24a}, which in contrast to ALTI-Logit and LRP does not consider the attention weights as a constant, and thus redistributes relevances backward onto the key and query vectors.
In particular \citet{pmlr-v235-achtibat24a} handles product layers by employing the LRP-uniform redistribution rule.
Concretely, given a product layer  $z_a \cdot z_b = z_j$, the relevance of the output neuron $R_j$ is redistributed equally among input neurons, hence $R_a = R_b = 0.5 \cdot R_j$.
This is similar to a rule previously proposed for extending LRP to customized LSTMs \cite{ArrXAI19,arjonamedina2019rudderreturndecompositiondelayed}.
As a result, the attention weights' matrix is assigned relevance scores,
opening up the question of how to redistribute this quantity further backward through the softmax non-linearity.
For that purpose \citet{pmlr-v235-achtibat24a} propose a novel redistribution rule which is equivalent to using the Gradient$\times$Input XAI-method for that layer.
While this redistribution strategy does not conserve the overall relevance between the layer's output and input neurons,
it can be justified by the fact that during the forward pass the softmax layer may have a non-zero output while all inputs are zero, which can be interpreted as a bias parameter for that layer\footnote{This is similar to how biases in linear layers get assigned (or absorb) a portion of the relevance. Indeed, strictly speaking, with LRP the sum of the input tokens' relevances will be numerically equal to the prediction logit only if all model biases are set to zero (which in practice can serve as a sanity check for the LRP implementation). See the redistribution rule for linear layers introduced in Section~\ref{sec:LRP}, where the bias term appears in the denominator.}.

Currently, an implementation of AttnLRP  \cite{pmlr-v235-achtibat24a} is available via the highly specialized LXT\footnote{\url{https://github.com/rachtibat/LRP-eXplains-Transformers/tree/25aa8f3}\\(latest available commit: Feb 13th 2025, 25aa8f3)} toolbox, which overwrites the Pytorch backward function of all layers present in the network. In Section~\ref{sec:LRPx}, we will show that a strategy similar to the one previously adopted for LRP based on a modified Gradient$\times$Input  approach can also be extended to AttnLRP to allow for a simpler and faster implementation. AttnLRP was evaluated against LRP from \citet{pmlr-v162-ali22a}, as well as various attention-based \cite{abnar-zuidema-2020-quantifying,9710570,9577970,deiseroth2023atman} and gradient-based \cite{Simonyan:ICLR2014,Sundararajan:ICML2017,Smilkov:ICML2017} XAI-methods, and AttnLRP was shown to deliver the best results.

\subsection{Gradient-based}

We consider two gradient-based methods commonly used in previous XAI works.
Both approaches compute the gradient of the prediction logit w.r.t.\ the input token's representation of interest and normalize it using either the $L_1$-norm or squared $L_2$-norm,
i.e., 
$R_t^0 = \|{\nabla}_{ x_t^0} \; \text{logit}_w\|{_1}$,
resp.\ $\|{\nabla}_{ x_t^0} \; \text{logit}_w\|{_2^2}$. Both variants have the advantage of being on an additive scale, meaning that the contributions of smaller units (neuron, token, or word) can be summed up to obtain the relevance of a greater portion of the input. We tried both and report only the best results under Gradient.
The Gradient$\times$Input method computes the dot product between the gradient and the input token's representation, i.e. $R_t^0 = {\nabla}_{ x_t^0} \cdot x_t^0 $.
All gradient-based methods are easy and efficient to compute, and can be obtained via standard gradient backpropagation.

\subsection{Overview}
Table~\ref{table:xai_methods_overview} summarizes all XAI attribution methods considered in this work. Only the first three methods ALTI-Logit, LRP and AttnLRP are decomposition-based and redistribute the prediction logit's quantity onto model components at different levels of granularity.
While ALTI-Logit assigns relevance at the token-level, and if desired also at the head-level inside MHA blocks,
LRP and AttnLRP are more fine-grained methods and decompose the prediction down to the smallest possible unit, i.e., a neuron.
Regarding computation time, all methods have conceptually a similar cost in number of forward/backward passes required, though depending on the efficiency of the particular implementation that is used different memory and time costs might arise in practice (as we will see for instance for AttnLRP in Section~\ref{sec:LRPx}).
In order to additively decompose the prediction logit into contributions of model components, decomposition-based XAI-methods make several simplifying assumptions: in particular they tend to "linearize" parts of the model (e.g., by viewing the attention matrix as a constant, or treating the standard deviation inside normalization layers as a constant, see Appendix~\ref{appendix:LRPx_proofs} for more details). Gradient-based explanations do not make those simplifications, though they are unable to explain the actual prediction's logit, but explain instead its derivatives (or in other words, per definition, they identify tokens/neurons of which a slight perturbation might influence a significant change in the prediction).
Finally, while most XAI-methods redistribute relevances backward across all layers of the model, and thereby take into account a mixing of contextual information arising from token-interactions inside MHA blocks, ALTI-Logit is the only method where the flow of information gets truncated inside MLP blocks and is not backward propagated further from these layers on (except for contributions from residual connections).

\begin{table*}
	\begin{center}
    \caption{Overview of the XAI attribution methods considered in this work.}
    \Large 
		\resizebox{\textwidth}{!}{
            \Large
			\begin{tabular}{l|cccccc}
            \toprule
             Method & \begin{tabular}{@{}c@{}}granularity \end{tabular} & \begin{tabular}{@{}c@{}}{computation}\end{tabular} & \begin{tabular}{@{}c@{}}treat normalization \\ as a linear layer\end{tabular} & \begin{tabular}{@{}c@{}}treat attention matrix \\ as a constant\end{tabular} & \begin{tabular}{@{}c@{}}mixing of information \\ upward MLP blocks\end{tabular}  & \begin{tabular}{@{}c@{}}logit  \\ decomposition\end{tabular} \\
            \midrule
            ALTI-Logit  & token, head, layer    & 2 $\times$ forward    &{\color{green} \ding{51}}  & {\color{green} \ding{51}}     & {\color{red} \ding{55}}       & {\color{green} \ding{51}}\\
            LRP         & neuron, layer         & forward $+$ backward  &{\color{green} \ding{51}}  & {\color{green} \ding{51}}     & {\color{green} \ding{51}}     & {\color{green} \ding{51}}\\
            AttnLRP     & neuron, layer         & forward $+$ backward  &{\color{green} \ding{51}}  & {\color{red} \ding{55}}       & {\color{green} \ding{51}}     & {\color{green} \ding{51}}\\
            \midrule
            \begin{tabular}{@{}l@{}} Gradient, \\ Gradient$\times$Input \end{tabular}   & neuron, layer         & forward $+$ backward  &{\color{red} \ding{55}}    & {\color{red} \ding{55}}       & {\color{green} \ding{51}}     & {\color{red} \ding{55}}\\
            \bottomrule
			\end{tabular}
		}
		\label{table:xai_methods_overview}
	\end{center}
\end{table*}

\subsection{Methods not considered}
Other non-decomposition based XAI-methods which we do not consider include more sophisticated gradient-based variants such as Integrated Gradients \cite{Sundararajan:ICML2017} and SmoothGrad \cite{Smilkov:ICML2017}. These methods try to alleviate the noisy gradient problem \cite{pmlr-v70-balduzzi17b} by averaging gradients over several perturbed samples. However they introduce hyperparameters into the explanation process (such as the number/type of perturbations or the baseline choice\footnote{Indeed these hyperparameters can have a huge impact on the quality of explanations, as was previously shown in computer vision \cite{ARRAS202214}: for instance a zero-valued baseline as is often used for the Integrated Gradients method might be sub-optimal.}), and in a typical XAI use-case with no available ground truth one has no criteria to tune those hyperparameters. Further, similarly to perturbation-based XAI-methods, generating and leveraging the perturbations yields an additional computation cost (one typically needs one backward, resp. forward, pass for each perturbed sample with gradient-based, resp. perturbation-based, XAI-methods).
Other popular non-decomposition based XAI-methods include attention-based methods such as Attention Rollout \cite{abnar-zuidema-2020-quantifying} and ALTI \cite{ferrando-etal-2022-measuring}. Although these methods are intuitively appealing since they leverage the mixing of information already provided by attention weights and trace it back across layers, those methods have been shown to be inferior to decomposition-based methods in previous works {\cite{pmlr-v162-ali22a,pmlr-v235-achtibat24a,ferrando-etal-2022-measuring,ferrando-etal-2023-explaining}}, and are typically not directly related to a specific class/token prediction.
Lastly, we do not consider other LRP-based approaches for Transformers proposed in the literature \cite{voita-etal-2021-analyzing,9577970}: those do not follow a layer-wise conservation principle within the MHA layer\footnote{In fact they enforce conservation artificially via a subsequent normalization step over relevances.} and have been observed to lead to numerical instabilities \cite{pmlr-v235-achtibat24a}.

\subsection{LRPx : Fast and simple implementation of LRP variants}\label{sec:LRPx}

The adoption of LRP, and its variant AttnLRP, has been so far mainly tied to the use of ready-made and highly specialized
toolboxes (such as Zennit \cite{anders2021software}, LXT \cite{pmlr-v235-achtibat24a} or others \cite{JMLR:v17:15-618,JMLR:v20:18-540}).
Such toolboxes compute the LRP relevances explicitly at each layer by overwriting the standard gradient backward pass (either through hooks, and by overwriting the backward function of every layer).
However, it is possible to implement LRP on Transformers in a more lightweight and elegant manner by adopting a \textit{modified} Gradient$\times$Input strategy. To the best of our knowledge the first work where this strategy was employed was \citet{eberle-etal-2023-rather}.
It consists in modifying a few layers during the forward pass (only non-linear layers need to be modified, so far less layers than in Zennit or LXT) such that their output values remain unchanged (hence without affecting the forward pass outcome), but in a way that the resulting gradients from the Pytorch´s automatic differentiation engine multiplied with the forward pass activations yields LRP relevances at any hidden or input layer of interest (in practice this is achieved by detaching dedicated neurons from the computational graph by using Pytorch's {\texttt{Tensor.detach()}} method).
Although this efficient and simple strategy to implement LRP has been further adopted in a recent work extending LRP to State Space Models \cite{NEURIPS2024_d6d0e41e}, and builds upon various LRP properties and implementation tricks provided in multiple previous works \cite{Montavon:ExplAIBook2020,thesis_Lapuschkin,thesis_Eberle,NEURIPS2024_d6d0e41e}, to the best of our knowledge there exist so far no comprehensive and complete set of proofs demonstrating the equivalence of explicit LRP rules with this \textit{modified} Gradient$\times$Input approach. In the present work we close this gap by providing such extensive proofs in the Appendix~\ref{appendix:LRPx_proofs}.

Further, we show for the first time that the \textit{modified} Gradient$\times$Input strategy can also be extended to AttnLRP.
As mentioned earlier, AttnLRP differs from LRP by the rules it employs for the product and softmax layers, see Section~\ref{sec:AttnLRP}. Let us consider the modified product layer defined by: $\hat{z}_j = 0.5 \cdot (z_a \cdot z_b) + {[0.5 \cdot (z_a \cdot z_b)]}_{\text{detach()}}$. One can easily see that the forward pass outcome remains unchanged (i.e., $z_j = \hat{z}_j)$. The resulting gradient of $z_a$ is: $d z_a = 0.5 \cdot z_b \cdot d z_j$. Now let's assume relevances are computed via a Gradient$\times$Input formula using this modified product layer, thus $R_j = d z_j \cdot z_j$ and $R_a = d z_a \cdot z_a$. As a result it holds: $R_a =  0.5 \cdot z_b \cdot d z_j \cdot z_a = 0.5 \cdot z_j \cdot d z_j = 0.5 \cdot R_j$, which is equivalent to the uniform rule for products $\square$.
Hence we have shown that the uniform rule used in AttnLRP can be implemented via a  \textit{modified} Gradient$\times$Input strategy. In the Appendix ~\ref{appendix:softmax} we provide a further proof that the AttnLRP redistribution rule for softmax proposed by \citet{pmlr-v235-achtibat24a} is equivalent to Gradient$\times$Input.

In this work we implement a straightforward and compact Pytorch toolbox named LRPx (where x stands for multiple LRP variants) which is part of our released code, and that allows to compute both LRP and AttnLRP using the \textit{modified} Gradient$\times$Input strategy. In Section~\ref{sec:speedup} we benchmark the resulting computational time on Transformers using LRPx against the LXT toolbox from \citet{pmlr-v235-achtibat24a}.

\FloatBarrier
\section{A Benchmark for Language Model Attributions} 

\subsection{Subject-Verb Agreement (SVA) Task} 
We build our XAI benchmark dataset for Language Models on top of the natural language subject-verb agreement dataset released by \citet{Goldberg_ArXiv2019}, which itself is based upon data from \citet{Linzen_TACL2016}. 
In order to identify the subject of a given verb  we employ Spacy's dependency parser and make sure that the dependency relation between the verb and the subject is of type ``nominal subject''.
Note that previous work by \citet{ferrando-etal-2023-explaining} also created a similar benchmark, however their ground truth subjects were incorrect\footnote{Indeed they used the first subject occuring in the sentence as ground truth, although it might not be in a dependency relation with the verb of interest in case of multi-phrase sentences. This bug has a huge impact on the results, e.g., on \texttt{GPT2-small} \citet{ferrando-etal-2023-explaining} report a MRR accuracy of approx. 0.60 for the ALTI-Logit method, while we find 0.81.}.
We build our dataset meticulously, additionally discarding some invalid and trivial samples, in order to release a proper and well-defined dataset to the research community.
Appendix~\ref{sec:appendix_benchmark_generation} provides all the details of the data generation process. Our resulting tokenized datasets contain 29k samples.

In order to explain the model's SVA predictions, we generate contrastive explanations \cite{yin-neubig-2022-interpreting}, in other words, we explain a logits difference of the form: $\text{logit}_{p} - \text{logit}_{o}$, where $p$ indicates the predicted verb number (singular/plural) and $o$ the opposite verb number. 

\subsection{Language Models}

We employ the following models: \texttt{bert-base-uncased}, \texttt{bert-large-uncased}, \texttt{GPT2-small}, \texttt{GPT2-XL}, \texttt{Llama-3.2-1B} and \texttt{Llama-3.2-3B} from the HuggingFace library. Appendix Table~\ref{table:prediction_accuracy} provides the models' prediction accuracy on SVA, as well as various informations on the models' sizes and tokenizer.

\subsection{Evaluation Metrics}
We employ four different evaluation metrics.

\textbf{{Pointing Game top-k} (PGk)}. This metric looks at the top-k tokens with the highest relevances. If one of these tokens is within the ground truth, the accuracy is 1 else 0. We report results for k=2 in our experiments. A similar metric has been previously used to evaluate attributions \cite{poerner-etal-2018-evaluating}.

\textbf{{Mean Reciprocal Rank} (MRR)}. This is the sole metric reported in the evaluation work by \citet{ferrando-etal-2023-explaining}. It consists in retrieving the inverse of the minimal rank (in decreasing order of relevance) of the tokens belonging to the ground truth.

\textbf{{Relevance Mass Accuracy} (RMA)}. This metric was introduced in computer vision \cite{ARRAS202214}, and calculates the fraction of positive relevance that falls inside the ground truth over the total positive relevance present in the input.

\textbf{{Per-Token Accuracy} (PTA)}. This metric makes a binary classification decision based on the sign of the relevance and then computes the classification accuracy w.r.t. the ground truth tokens. More precisely, it assumes tokens inside the ground truth shall receive a strictly positive relevance, while tokens outside the ground truth shall have no relevance or a negative relevance. It is related to the Pixel Accuracy used to evaluate semantic segmentation in computer vision.

\section{Results}

\subsection{Evaluation w.r.t. Ground Truth} 

Table~\ref{table:relevance_accuracy} presents our results. We computed the evaluation metrics using only correctly predicted samples\footnote{We generated the token-level attributions using double precision, except for \texttt{Llama-3.2-3B}, where we use half precision \texttt{bfloat16} (though our experiments revealed that the precision has almost no impact on the metrics' results).}.
When we look at the PTA results, we find that Gradient has the best performance, and results are consistent across models. However, the score achieved by Gradient for this metric is close to random, and worse than random for the other XAI-methods. This illustrates the importance of the choice of the metric for XAI evaluation, and reveals that the underlying assumption of PTA that the sign of the relevance shall switch between tokens inside and outside the ground truth is not adequate.

The remaining evaluation metrics deliver largely consistent result per Language Model family. While AttnLRP performs well on BERT and Llama3 models, ALTI-Logit is the strongest method on GPT2, followed by LRP.
We are not yet able to understand why the results are so different across model families, since most components within the given architectures are similar. However we believe this constitutes an interesting finding that should be investigated in future work.
In terms of magnitude of the metrics, results are higher for GPT2 and Llama3 than on BERT, which is probably due to the different tokenizers and vocabulary sizes that lead to longer inputs for BERT models, since this difference is also reflected in the random baseline results.

\begin{table*}
	\begin{center}
            \caption{Token-level accuracy of the XAI-methods w.r.t. ground truth, using different metrics (PG2: Pointing Game top-2, MRR: Mean Reciprocal Rank, RMA: Relevance Mass Accuracy, PTA: Per-Token Accuracy). All metrics are within [0.0, 1.0], the higher the better, we highlight in bold the best result, and underline the second best per model. The random baseline was obtained by sampling relevances uniformly in the range [-1.0, 1.0) for each given model's tokenized dataset (averaged over 10 runs).}
    \small 
		\resizebox{1.0\textwidth}{!}{
			\begin{tabular}{l|cccc|cccc|cccc}
            \toprule
            {} & \multicolumn{4}{c}{\normalsize BERT} & \multicolumn{4}{c}{\normalsize  GPT2} & \multicolumn{4}{c}{\normalsize Llama3} \\[4pt]
            {} & \multicolumn{4}{c}{\texttt{bert-base-uncased}} & \multicolumn{4}{c}{\texttt{GPT2-small}} & \multicolumn{4}{c}{\texttt{Llama-3.2-1B}}  \\[5pt]  
            XAI-Method & PG2$\uparrow$ & MRR$\uparrow$ &  RMA$\uparrow$  & PTA$\uparrow$ & PG2 & MRR &  RMA  & PTA & PG2 & MRR &  RMA  & PTA \\[3pt]
			\midrule
            ALTI-Logit              & - & - & - & -                                                                         & \textbf{0.867}    & \textbf{0.808}    & \underline{0.342}    &  0.330                & \underline{0.690}  & \underline{0.623} &  \underline{0.288}    & 0.359 \\  
            LRP                     & 0.688             & 0.637             &      0.208                & 0.339             & \underline{0.738} & \underline{0.706} & \textbf{0.367}       &  \underline{0.445}    & \underline{0.690}  & 0.533             &  0.221                & 0.244 \\  
            AttnLRP                 & \textbf{0.775}    & \underline{0.705} &      \textbf{0.260}       & \underline{0.376} & 0.705             & 0.634             & 0.318                &  0.408                & \textbf{0.884}     & \textbf{0.814}    &  \textbf{0.387}       & \underline{0.382} \\ 
            Gradient                & \textbf{0.775}    & \textbf{0.718}    &      \underline{0.245}    & 0.041             & 0.592             & 0.551             & 0.255                &  0.153                & 0.283              & 0.366             &  0.151                & 0.127 \\ 
            Gradient$\times$Input   & 0.292             & 0.316             &      0.095                & \textbf{0.497}    & 0.262             & 0.353             & 0.152                &  \textbf{0.565}       & 0.365              & 0.407             &  0.183                & \textbf{0.512} \\[5pt]
            {} & \multicolumn{4}{c}{\texttt{bert-large-uncased}} & \multicolumn{4}{c}{\texttt{GPT2-XL}} & \multicolumn{4}{c}{\texttt{Llama-3.2-3B}} \\[3pt]
            \midrule
            ALTI-Logit              & - & - & - & -                                                                         & \textbf{0.885}    & \textbf{0.852}    &  \underline{0.402}    & 0.320                & 0.614              & 0.557              &  \underline{0.266}   & 0.297\\
            LRP                     & \underline{0.560} & 0.521             &      \underline{0.187}    & \underline{0.408} & \underline{0.823} & \underline{0.754} &  \textbf{0.407}       & \underline{0.392}    & \underline{0.747}  & \underline{0.622}  &  0.241               & 0.231 \\
            AttnLRP                 & \textbf{0.641}    & \textbf{0.580}    &      \textbf{0.224}       & 0.383             & 0.779             & 0.658             &  0.351                & {0.384}              & \textbf{0.885}     & \textbf{0.764}     &  \textbf{0.364}      & \underline{0.326} \\ 
            Gradient                & 0.555             & \underline{0.551} &      0.177                & 0.041             & 0.579             & 0.546             &  0.265                & 0.153                & 0.275              & 0.311              &  0.143               & 0.127 \\ 
            Gradient$\times$Input   & 0.212             & 0.249             &      0.077                & \textbf{0.513}    & 0.321             & 0.408             &  0.198                & \textbf{0.608}       & 0.377              & 0.413              &  0.192               & \textbf{0.514} \\[1pt]
            \midrule
            Random \; {\scriptsize mean}                & 0.080 & 0.149 & 0.040 & 0.500    & 0.277                     &  0.360            &  0.151            & 0.501                & 0.241             & 0.326              &  0.127               & 0.501 \\
            \hphantom{Random} \; {\scriptsize$\pm$std} & 0.002 & 0.001 & 0.000 & 0.001    & 0.004                     & 0.002             &  0.001            & 0.001                & 0.002             & 0.001              &  0.001               & 0.000 \\
            \bottomrule
			\end{tabular}
		}
		\label{table:relevance_accuracy}
	\end{center}
\end{table*}

\subsection{Exemplary Heatmaps}

In the Appendix Fig.~\ref{heatmap} we provide some exemplary heatmaps using the five samples with the highest predicted logits difference across the dataset for the model {\texttt{Llama-3.2-1B}. One can see that heatmaps for AttnLRP are more sparse and focused on the ground truth subject than those for ALTI-Logit. While ALTI-Logit also assigns a high relevance to the subject, it additionally gives a high importance to the special token \texttt{<begin\_of\_text>}. Heatmaps for Gradient$\times$Input and Gradient on the other hand look very noisy.

\subsection{Computational Speedup with LRPx}\label{sec:speedup}

We calculated the computational time speedup obtained for AttnLRP using our LRPx toolbox (i.e., a Gradient$\times$Input strategy), versus the original LXT toolbox from \citet{pmlr-v235-achtibat24a} (based on an explicit relevance computation). To this end, we retrieve the median speedup over the first 1,000 samples of our SVA dataset, using two types of GPUs: NVIDIA Tesla V100 32GB and NVIDIA A100 40GB, and single precision. On \texttt{bert-base-uncased} we obtained a speedup between 1.83 and 1.98, and on \texttt{Llama-3.2-1B} between 1.54 and 1.61. Hence this illlustrates that the Gradient$\times$Input approach for implementing LRP/AttnLRP is not only conceptually simpler, but also faster.

\section{Outlook}

While in this work we have focused on evaluating decomposition-based attributions on the input tokens, since for the input tokens one can easily define a ground truth, the relevances obtained for hidden layers might in principle also be useful to perform other tasks than merely explaining the predictions, e.g., in order to unbias or improve the model's performance \cite{WEBER2023154}, increase model robustness to perturbations \cite{sun2025transformerlayerspainters}, or to prune and quantize the model \cite{YEOM2021107899,Becking2022}. 
Besides it has been recently shown that gradient-based relevances can be used in place of costly causal attribution methods to localize and control model behaviors to components \cite{kramár2024atpefficientscalablemethod}. The decomposition-based approaches discussed in this work might perform even better in this regard, since their token-level accuracies are generally higher than those of gradient-based methods.
Another complementary direction to the present evaluation approach would be to consider synthetic tasks to evaluate XAI \cite{bastings-etal-2022-will}, in order to allow for a better control over biases and Clever Hans behaviors \cite{Lap:Nature19}, or to use white-box models \cite{hao-2020-evaluating}.
Lastly, our evaluation approach using subject-verb agreement on Transformers can also be extended to more recent Language Model architectures, such as State Space Models \cite{Gu2021EfficientlyML} and xLSTMs \cite{NEURIPS2024_c2ce2f27}.

\FloatBarrier
\section{Conclusion}
In this work we took a close look at state-of-the-art decomposition-based attribution methods by analyzing their common properties, as well as their differences. Further we showed that LRP, as well as AttnLRP, can be computed in a simple and fast way by using 
a \textit{modified} Gradient$\times$Input strategy. Our careful evaluation w.r.t. automatically generated ground truth annotations reveals that the quality of explanations differs across model families. Identifying the root causes for these differences shall constitute a topic for future work.

\section*{Limitations}
Our ground truth annotations were automatically generated using the Spacy's dependency parser. Such a a parser is not 100\% accurate, and hence might introduce some noise in the evaluation process. Further, our benchmark dataset is extracted from real-world natural language data, and as such it might contain misspellings, typographical errors, and even grammatically incorrect sentences. However our goal in this work is to evaluate XAI in a realistic setup, and we believe those limitations do not influence the relative comparison of XAI-methods in a noticeable way.

\section*{Acknowledgments}

This work was supported by the Federal Ministry of Education and Research (BMBF) as grant BIFOLD (01IS18025A, 01IS180371I); the European Union’s Horizon Europe research and innovation programme (EU Horizon Europe) as grants [ACHILLES (101189689), TEMA (101093003)]; and the German Research Foundation (DFG) as research unit DeSBi [KI-FOR 5363] (459422098).



\bibliography{bibliography}

\appendix
\newpage
\section{Software Requirements \& Licenses}
\label{sec:appendix_software_requirements}
All our experiments are conducted using the following python packages and their respective version numbers within a Python 3.11.9 environment:
\begin{itemize}
  \item Spacy 3.7.3
  \item Pandas 2.2.1
  \item Pytorch 2.3.0
  \item Numpy 1.26.4
  \item HuggingFace Transformers 4.48.1
\end{itemize}

\paragraph{\textbf{Licenses}} BERT is released under Apache 2.0, GPT-2 under MIT, and Llama-3 under Meta Llama 3 Community License. 

\section{XAI Benchmark Dataset Generation}
\label{sec:appendix_benchmark_generation}

We build our XAI benchmark for language models on top of the natural language subject-verb agreement dataset released by \citet{Goldberg_ArXiv2019} (available under: \url{https://github.com/yoavg/bert-syntax/blob/master/lgd_dataset.tsv}), which itself is based upon data from \citet{Linzen_TACL2016} with MIT license. This dataset is made of initially 29,985 uncased sentences from Wikipedia, each containing a verb in present tense, and allowing for a bidirectional stimuli with input beyond the verb's position in the sentence (i.e., for BERT-like masked language models). For causal language models (i.e., GPT2-like language models) we use as a stimuli only the portion of the sentence \textit{before} the verb's position. Each sentence additionally contains at least one agreement ``attractor'' located between the subject and the verb (the number of attractors per sample varies between 1 and 4), and all attractors are nouns of opposite number from the subject, which makes this dataset well-suited for XAI evaluation, as the evidence for the correct verb number shall be concentrated on the subject. We noticed that the original dataset from \citet{Goldberg_ArXiv2019} contained 46 invalid samples, where the singular and plural verb forms were identical, which we discarded from our benchmark. 

In the following we describe how we identify the subject of each sentence (i.e. the linguistic evidence we use as the ground truth for the XAI evaluation), as well as the preprocessing steps we undertook to take into account each language model's specific tokenization.

\textbf{Generic ground truth.} In a first step we generate the model-agnostic ground truth data. For that purpose we use Spacy's dependency parser from the english pipeline \texttt{en\_core\_web\_trf} to identify the subject of a given verb in a sentence. We retain only samples with the syntactic dependency relation \texttt{nsubj} (i.e., ``nominal subject''), thereby we aim to remove potential ambiguous cases (this step discards 1005 samples). Further, we retain only samples whose verb was identified by Spacy's part-of-speech tagger to be either of type \texttt{VBZ} (``verb, 3rd person singular present'') or \texttt{VBP} (``verb, non-3rd person singular present'') (thereby discarding 148 samples where the verb was not recognized as being conjugated in present tense). This leaves us with a dataset of size 28,786, from which 67\% of the samples contain a ``plural'' verb form as the correct prediction (note that such ``plural'' verb forms also include some rare samples where the pronouns ``I'' and ``you'' are the subject, and thus strictly speaking would be singular cases), and 33\% of the samples contain a ``singular'' verb form, hence the verb's number is imbalanced.

\textbf{Tokenized ground truth.} For each considered language model, we generate in a second step a model-specific benchmark made of tokenized stimulis and their corresponding tokenized ground truths, by taking into account each model's particular tokenizer. More precisely, we discard samples for which the verb's singular or plural inflection gets tokenized into more than one token, since
the SVA prediction is based on comparing the logit scores for these two verb forms. Further, we verify that the ground truth is always shorter (in terms of number of tokens) than the input text stimuli to avoid any trivial cases for XAI evaluation. For causal language models (i.e. GPT2-like), we also discard samples where a portion of the ground truth lies \textit{after} the verb in the sentence. Finally, we ensure that the \textit{effective} input text (i.e. when excluding some special tokens, such as \texttt{[CLS], [SEP] and [MASK]} for BERT) is always longer than one token, again to avoid any trivial cases for XAI evaluation.

With the above considerations, we finally obtain for BERT a benchmark made of 28472 samples, whose input length (in terms of number of tokens) varies between 9 and 170, with mean 30, std 12, and median of 28, while the ground truth's length varies between 1 and 7 with 96.7\% of the samples having a ground truth length of 1 (and 2.6\% of samples a ground truth length of 2).

For GPT-2 we likewise obtain a benchmark made of 28,602 samples, whose input length (in terms of number of tokens) varies between 2 and 60, with mean 11, std 7, and median of 8, while the ground truth's length varies between 1 and 7 with 85.1\% of the samples having a ground truth length of 1 (and 12.4\% of samples a ground truth length of 2). 

For Llama-3 we finally obtain a benchmark made of 28,629 samples, with an input length (in number of tokens) varying between 3 and 60, with mean 11, std 7, and median of 9, while the ground truth's length varies between 1 and 5 with 89.4\% of the samples having a ground truth length of 1 (and 8.8\% of samples a ground truth length of 2).

\section{Language Models}

Table~\ref{table:prediction_accuracy} summarizes the prediction accuracy of each model on our tokenized subject-verb agreement benchmark datasets, as well as provides various informations about the models' sizes and tokenizers.

\begin{table*}
	\begin{center}
    \caption{Prediction accuracy on subject-verb agreement, and model information.}
		\resizebox{\textwidth}{!}{
			\begin{tabular}{l|ccccccc}
            \toprule
            Model                           & prediction accuracy   & \# params     & \# layers     & \# heads  & hidden size       & vocab size        & tokenizer \\
            \midrule
            {\texttt{bert-base-uncased}}    & 0.969                     & 110M             & 12             & 12         & 768               & 30522             & WordPiece \\
            {\texttt{bert-large-uncased}}   & 0.974                     & 340M             & 24             & 16         & 1024              & same              & same \\
            \midrule
            {\texttt{GPT2-small}}           & 0.919                     & 124M             & 12             & 12         & 768               & 50257             & BPE \\
            {\texttt{GPT2-XL}}              & 0.941                     & 1.5B             & 48             & 25         & 1600              & same              & same \\
            \midrule
            {\texttt{Llama-3.2-1B}}          & 0.954                     & 1B             & 16             & 32         & 2048              & 128256            & tiktoken BPE \\
            {\texttt{Llama-3.2-3B}}          & 0.956                     & 3B             & 28             & 24         & 3072              & same              & same \\
            \bottomrule
			\end{tabular}
		}
		\label{table:prediction_accuracy}
	\end{center}
\end{table*}

\FloatBarrier
\section{Proofs on implementing LRP variants via a \textit{modified} Gradient$\times$Input strategy}\label{appendix:LRPx_proofs}

In this Section we build upon various derivations of LRP rules' properties and implementation tricks employed in previous works \cite{Montavon:ExplAIBook2020,thesis_Lapuschkin,thesis_Eberle,NEURIPS2024_d6d0e41e}, in order to provide a unified and complete set of proofs demonstrating that the LRP explanation method, and its variants, can be implemented in a
simple and elegant manner using a \textit{modified} Gradient$\times$Input strategy.

\subsection{LRP-$\epsilon$ rule for linear layers}\label{appendix:linear_layer}

Given a linear layer of the form $ z_j = \sum_i z_i w_{ij} + b_j\;$ in the forward pass, and given the relevances of the output neurons $R_j$,  
the input neurons' relevances $R_i$ are computed using the following LRP-$\epsilon$ rule:
\begin{equation}
R_{i} = \sum_j \tfrac{z_i \cdot w_{ij}}{z_j  \;  + \; \epsilon \cdot {\text{sign}}(z_j)} \cdot R_j \,
\end{equation}
where the term $\epsilon$ is a small positive numerical stabilizer. But for simplifying the derivation let's assume $\epsilon=0$, and so: $ R_{i} = \sum_j \tfrac{z_i \cdot w_{ij}}{z_j} \cdot R_j \,$.

Now let's assume the relevances at the layer output and input are computed via Gradient$\times$Input, in other words it holds: 
\begin{align}
R_j=d z_j \cdot z_j \label{eq:linear_GI_1}\\
R_i=d z_i \cdot z_i  \label{eq:linear_GI_2}
\end{align}
Using elementary rules of differentiation and the chain rule it holds: 
\begin{equation}
d z_i = \sum_{j} d z_j \cdot w_{ij} \label{eq:linear_chain}
\end{equation}
By incorporating Eq.~\ref{eq:linear_chain} into Eq.~\ref{eq:linear_GI_2}, we obtain:
\begin{equation}
R_i= z_i \cdot \sum_{j} d z_j \cdot w_{ij}
\end{equation}
And then replacing $d z_j$ by its value from Eq.~\ref{eq:linear_GI_1}, we finally get:
\begin{equation}
R_i= z_i \cdot \sum_{j} \tfrac{R_j}{z_j} \cdot w_{ij}
\end{equation}
And by rearranging terms:
\begin{equation}
R_{i} = \sum_j \tfrac{z_i \cdot w_{ij}}{z_j} \cdot R_j \;\;\; \square
\end{equation}
Hence we have shown that using Gradient$\times$Input one can implement the LRP-$\epsilon$ rule with $\epsilon=0$.
Using the Gradient$\times$Input strategy presents even an advantage over an explicit implementation of the LRP-$\epsilon$ rule. Indeed with Gradient$\times$Input no fraction is involved in the computation, and hence no denominator needs to be stabilized,
while with explicit LRP one has to use a non-zero $\epsilon$ stabilizer, which might introduces some noise or dampen the explanation process, as the $\epsilon$ value is kind of arbitrary, and its impact will be higher the lower the magnitude of the denominator`s value.

\subsection{LRP-$\alpha,\beta$ rule for linear layers}

While in the NLP domain the LRP-$\epsilon$ rule previously introduced has been shown to work well \cite{pmlr-v235-achtibat24a,arras-etal-2019-evaluating,Arras:PLOSONE2017,poerner-etal-2018-evaluating}, in the computer vision domain, in particular on convolutional layers, other rules such as the LRP-$\alpha,\beta$ rule and LRP-$\gamma$ rule have been shown to be more adequate \cite{Montavon:ExplAIBook2020,ARRAS202214,9206975}.
As an  example and for the sake of completeness, we demonstrate that the LRP-$\alpha_1,\beta_0$ rule can also be implemented via a \textit{modified} Gradient$\times$Input strategy. One advantage of the LRP-$\alpha_1,\beta_0$ rule over the LRP-$\gamma$ rule and the general LRP-$\alpha,\beta$ rule is that LRP-$\alpha_1,\beta_0$ has no free hyperparameter.

Given a linear layer of the form $ z_j = \sum_i z_i w_{ij} + b_j\;$ in the forward pass, and given the relevances of the output neurons $R_j$,  
using the LRP-$\alpha_1,\beta_0$ rule, the input neurons' relevances $R_i$ are computed as:
\begin{equation}
R_{i} = \sum_j \tfrac{(z_i \cdot w_{ij})^+}{z_j^{\text{pos}}} \cdot R_j \,
\end{equation}
where ${z_j^{\text{pos}}} = \sum_i (z_i w_{ij})^+ + (b_j)^+$ and $(\cdot)^+ = \text{max} (0, \cdot)$. 

Now let's define a modified forward function of the layer of the form:
\begin{equation}
\hat{z}_j = z_j^{\text{pos}} \cdot {[ \tfrac{z_j}{z_j^{\text{pos}}}]}_{\text{detach()}} \label{eq:alpha1_0}
\end{equation}
One can see that this modification does not affect the forward pass outcome, in other words it holds: $\hat{z}_j = z_j$.

Now assuming that the relevances at the layer's output and input are computed via Gradient$\times$Input, thus:
\begin{align}
R_j=d \hat{z}_j \cdot \hat{z}_j \label{eq:alpha1_1}\\
R_i=d z_i \cdot z_i  \label{eq:alpha1_2}
\end{align}

Using elementary rules of differentiation and the chain rule on the definition of $z_j^{\text{pos}}$, we obtain:
\begin{align}
d{z_i} &= \sum_j 
                \begin{cases}
                (w_{ij})^+ \cdot dz_j^{\text{pos}}          & \text{for } z_i \geq 0 \\
                (w_{ij})^- \cdot dz_j^{\text{pos}}          & \text{for } z_i < 0 
                \end{cases}   \label{eq:alpha1_3}
\end{align}
where we use the notation $(\cdot)^- = \text{min} (0, \cdot)$.

By incorporating Eq.~\ref{eq:alpha1_3} into Eq.~\ref{eq:alpha1_2}, we get:
\begin{align}
R_i &= \sum_j 
                \begin{cases}
                (w_{ij})^+ \cdot z_i \cdot dz_j^{\text{pos}}          & \text{for } z_i \geq 0 \\
                (w_{ij})^- \cdot z_i \cdot dz_j^{\text{pos}}          & \text{for } z_i < 0 
                \end{cases}   \label{eq:alpha1_4} \\
    &= \sum_j 
                (z_i w_{ij})^+ \cdot dz_j^{\text{pos}} \label{eq:alpha1_5}
\end{align}

Further, using an elementary rule of differentiation on Eq.\ref{eq:alpha1_0}, and the equivalence $\hat{z}_j = z_j$, we have:
\begin{equation}
dz_j^{\text{pos}} = { \tfrac{z_j}{z_j^{\text{pos}}}} \cdot d\hat{z}_j = \tfrac{d\hat{z}_j  \cdot \hat{z}_j}{z_j^{\text{pos}}}  \label{eq:alpha1_6}
\end{equation}

By incorporating Eq.~\ref{eq:alpha1_1} into Eq.~\ref{eq:alpha1_6}, we get:
\begin{equation}
dz_j^{\text{pos}} = { \tfrac{R_j}{z_j^{\text{pos}}}}   \label{eq:alpha1_7}
\end{equation}

And by incorporating the latter into Eq.\ref{eq:alpha1_5}, we finally arrive at the LRP-$\alpha_1,\beta_0$ rule:
\begin{equation}
R_{i} = \sum_j \tfrac{(z_i \cdot w_{ij})^+}{z_j^{\text{pos}}} \cdot R_j \,  \;\;\; \square
\end{equation}

And by the way this rule does not need any numerical stabilizer. Indeed in the particular case where the denominator in the fraction $\tfrac{(z_i \cdot w_{ij})^+}{z_j^{\text{pos}}}$ is zero, one can be sure that the numerator is also zero, which can be handled by replacing the whole fraction with zero in this case, hence no relevance is propagated backward for such an output neuron $z_j$. With the \textit{modified} Gradient$\times$Input strategy, this case is handled similarly by setting: $\hat{z}_j = [z_j]_{\text{detach()}}$ if $z_j^{\text{pos}}=0$. This way the forward pass outcome remains unaffected, but the resulting gradient will be zero, again assigning no relevance backward from such a neuron.

\subsection{LRP-identity rule for element-wise activation layers}

Given an element-wise activation layer of the form: $z_j = g(z_i)$, with $g$ being the activation function. The LRP-identity rule redistributes the relevance identically from the layer's output to the layer's input, thus $R_i = R_j$.

Now let's define a modified forward function for the layer of the form: 
\begin{equation}
\hat{z}_j = z_i \cdot {[ \tfrac{g(z_i)}{z_i}]}_{\text{detach()}}
\end{equation}

Obviously it holds that $\hat{z}_j = z_j$, so the forward pass outcome remains unchanged.

Using elementary rules of differentiation and the chain rule it holds: 
\begin{equation}
d z_i = {[ \tfrac{g(z_i)}{z_i}]} \cdot d \hat{z}_j
\end{equation}

Now assuming we compute the relevances at the layer's output as well as at the layer's input with Gradient$\times$Input using the modified layer, we get:
\begin{align}
R_i &= d {z}_i \cdot z_i = {[ \tfrac{g(z_i)}{z_i}]} \cdot d \hat{z}_j \cdot z_i \\
    &= g(z_i) \cdot d \hat{z}_j = \hat{z}_j \cdot d \hat{z}_j \\
    &= R_j  \;\;\; \square
\end{align}
Hence we have shown the LRP-identity rule can be implemented implicitly by using the \textit{modified} Gradient$\times$Input strategy.

Moreover, note that one does not even need a numerical stabilizer to handle a zero-valued input in the activation layer. Indeed  most considered element-wise activation functions (such as GELU or SiLU) have a zero-valued output when their input is zero. 
Thus one possibility to deal with a zero-valued input is to set the output manually to zero for $\hat{z}_j$ in this particular case (i.e., to a constant), hence the resulting gradient will be zero. And as a consequence, the relevance using the Gradient$\times$Input strategy will be zero.
This is still meaningful for LRP as in such a case the output's relevance will be zero anyway, so there is no relevance to redistribute backward (indeed LRP relevances are generally proportional to neurons' contributions in the forward pass, and for a subsequent linear layer a zero-valued input does not contribute to the output, hence receiving no relevance).

\subsection{LRP-signal-take-all rule for product layers}

Given a product layer of the following form: $z_j = z_g \cdot z_s$, where $z_g$ is a gate neuron and $z_s$ is a signal neuron (in the MHA attention layer the former will be the attention weight, and the latter a component of the value vector).

The LRP-signal-take-all rule redistributes all the relevance to the signal neuron, i.e., $R_s = R_j$ and $R_g = 0$.

And let's define the following modified product layer: 
\begin{equation}
\hat{z}_j = {[z_g]}_{\text{detach()}} \cdot z_s \label{eq:product_0}
\end{equation}
Obviously $\hat{z}_j = z_j$.

Now assuming we compute the relevances at the layer's output and input via Gradient$\times$Input, thus it holds:
\begin{align}
R_j=d \hat{z}_j \cdot \hat{z}_j \label{eq:product_1}\\
R_s=d z_s \cdot z_s  \label{eq:product_2}\\
R_g=d z_g \cdot z_g  \label{eq:product_3}
\end{align}

Per definition of elementary rules of differentiation and the chain rule on Eq.\ref{eq:product_0}, we have:
\begin{align}
d z_s &= d \hat{z}_j \cdot z_g  \label{eq:d_1}\\
d z_g &= 0  \label{eq:d_2}
\end{align}

By incorporating Eq.~\ref{eq:d_1}\&\ref{eq:d_2} into Eq.~\ref{eq:product_2}\&\ref{eq:product_3}, we finally get:
\begin{align}
R_s &= d \hat{z}_j \cdot z_g \cdot z_s = d \hat{z}_j \cdot \hat{z}_j  \label{eq:r_1}\\
R_g &= 0 \label{eq:r_2}
\end{align}

And by using Eq.~\ref{eq:product_1}:
\begin{align}
R_s &= R_j   \\
R_g &= 0   \;\;\; \square
\end{align}
Hence we have shown the LRP-signal-take-all rule  can be implemented implicitly by using the \textit{modified} Gradient$\times$Input strategy.

\subsection{LRP rule for normalization layers}

We illustrate this rule using the Pytorch LayerNorm layer (a similar rule can be applied to RMSNorm), which is defined by:

\begin{equation}
z_j = \tfrac{z_i - E[z_i]}{\sqrt{Var[z_i] + \epsilon}} \cdot \gamma + \beta
\end{equation}
where the parameters of the layers $\epsilon$, $\gamma$ and $\beta$ are constants.

In order to extend LRP to Transformers \citet{pmlr-v162-ali22a} propose to treat the standard deviation of the Layernorm as a constant, which can be achieved by modifying the layer in the following way:
\begin{equation}
\hat{z}_j = \tfrac{z_i - E[z_i]}{{[\sqrt{Var[z_i] + \epsilon}]}_\text{detach()}} \cdot \gamma + \beta
\end{equation}
Obviously $\hat{z}_j = z_j$.

Further the modified layer is now a linear layer (since all remaining operations in the layer such as the mean operation are linear).
Hence the layer can be treated similarly to Section~\ref{appendix:linear_layer}.    \;\;\; $\square$

So overall we have shown that the LRP rule proposed for normalization layers in Transformers can be implemented with Gradient$\times$Input.

\subsection{AttnLRP rule for softmax layers}\label{appendix:softmax}

Let us introduce some new notations to match closely the ones from \citet{pmlr-v235-achtibat24a}. So far we have mainly dealt with single neurons, now we deal with vectors.
So let the input vector be $\mathbf{x}$ and the output vector be $\mathbf{s}$, and both can be indexed either by $i$ or $j$.

Per defintion of the softmax operation we have:
\begin{equation}
s_j (\mathbf{x}) = \tfrac{e^{x_j}}{\sum_i e^{x_i}}
\end{equation}

Using elementary rules of differentiation one can show that:
\begin{equation}
    \frac{\partial s_j}{\partial x_i} = 
    \begin{cases}
      s_j(1-s_j) & \text{for } i=j \\
        -s_j s_i & \text{for } i \neq j
    \end{cases} \label{eq:d_softmax}
\end{equation}

Now assuming the input's and output's relevances are computed via Gradient$\times$Input, i.e.:
\begin{align}
R_{s_j} &= d s_j \cdot s_j   \label{eq:identify} \\
R_{x_i} &= d x_i \cdot x_i 
\end{align}

Using the chain rule it holds that:
\begin{align}
R_{x_i} &= d x_i \cdot x_i \\
        &= \sum_j \frac{\partial s_j}{\partial x_i}  \cdot d s_j \cdot x_i \label{eq:R_x_i}
\end{align}

By incorporating Eq.~\ref{eq:d_softmax} into Eq.~\ref{eq:R_x_i}, one obtains:
\begin{align}
R_{x_i} &= \sum_j 
                \begin{cases}
                s_j(1-s_j)  \cdot d s_j \cdot x_i                 & \text{for } i=j \\
                -s_j s_i    \cdot d s_j \cdot x_i                 & \text{for } i \neq j
                \end{cases} \label{eq:softmax_2}
\end{align}

By identifying the term from Eq.~\ref{eq:identify}:
\begin{align}
R_{x_i} &= \sum_j 
                \begin{cases}
                (x_i-s_j\cdot x_i )      \cdot   R_{s_j}          & \text{for } i=j \\
                - s_i     \cdot x_i     \cdot   R_{s_j}         & \text{for } i \neq j
                \end{cases}   
\end{align}
Hence we finally arrived at the LRP rule proposed for the softmax layer in \citet{pmlr-v235-achtibat24a} (see Appendix A.3.1 of their work). \; $\square$

Thus, in summary, we have provided a complete set of proofs that all LRP, resp. AttnLRP, rules used in Transformers can be implemented via a
Gradient$\times$Input approach, by simply modifying adequately parts of the non-linear layers (namely product, normalization and element-wise activation layers) and keeping all linear layers unmodified.

\begin{figure*}[ht]
    \centering
    \includegraphics[trim={0cm 0cm 0cm 0cm}, clip, scale=0.36]{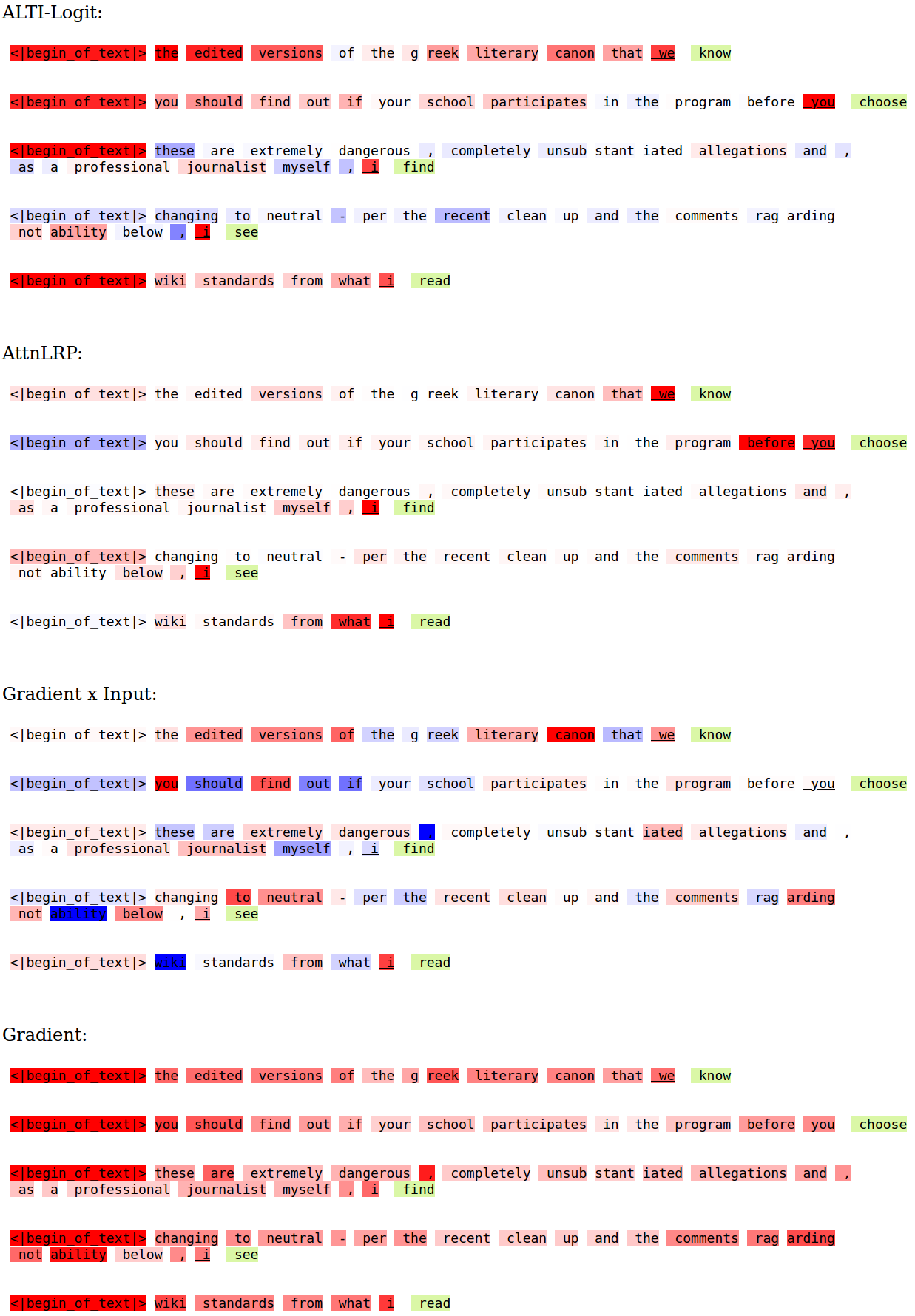}
    \caption{Exemplary heatmaps on correctly predicted samples for the \texttt{Llama-3.2-1B} model. The predicted verb in highlighted in green, positive relevance is mapped to red, negative to blue. The ground truth subject is underlined (in all considered samples it is the token preceding the verb). }
    \label{fig:benchmarkpipeline}
\end{figure*}\label{heatmap}

\end{document}